\documentclass[manuscript,screen]{acmart}

\AtBeginDocument{%
  }

\setcopyright{acmlicensed}
\copyrightyear{2025}
\acmYear{2025}
\acmDOI{XXXXXXX.XXXXXXX}

\acmISBN{978-1-4503-XXXX-X/2018/06}

\usepackage{multirow}
\usepackage{subfigure}
\usepackage{graphicx}
\usepackage{array}
\usepackage{amsmath}
\usepackage{algorithmic}
\usepackage{color}
\usepackage{url}

\begin{document}

\title{Visual Self-paced Iterative Learning for Unsupervised Temporal Action Localization}

\author{Yupeng~Hu}
\affiliation{%
  \institution{Shandong University}
  \city{Jinan}
  \state{Shandong}
  \country{China}
}
\email{huyupeng@sdu.edu.cn}

\author{Han Jiang}
\affiliation{%
  \institution{Xi’an Jiaotong University}
  \city{Xi’an}
  \state{Shaanxi}
  \country{China}
}
\email{jh.lumen@gmail.com}

\author{Hao~Liu}
\affiliation{%
  \institution{Shandong University}
  \city{Jinan}
  \state{Shandong}
  \country{China}
}
\email{liuh90210@gmail.com}

\author{Kun~Wang}
\affiliation{%
  \institution{Shandong University}
  \city{Jinan}
  \state{Shandong}
  \country{China}
}
\email{khylon.kun.wang@gmail.com}

\author{Haoyu Tang}
\affiliation{%
  \institution{Shandong University}
  \city{Jinan}
  \state{Shandong}
  \country{China}
}
\email{tanghao258@sdu.edu.cn}
\authornote{Haoyu Tang is the corresponding author.}

\author{Liqiang~Nie}
\affiliation{%
  \institution{Harbin Institute of Technology (Shenzhen)}
  \city{Shenzhen}
  \state{Guangdong}
  \country{China}
}
\email{nieliqiang@gmail.com}

\renewcommand{\shortauthors}{Hu et al.}

\begin{abstract}
  Recently, temporal action localization (TAL) has garnered significant interest in information retrieval community. However, existing supervised/weakly supervised methods are heavily dependent on extensive labeled temporal boundaries and action categories, which is labor-intensive and time-consuming. Although some unsupervised methods have utilized the ``iteratively clustering and localization'' paradigm for TAL, they still suffer from two pivotal impediments: 1) unsatisfactory video clustering confidence, and 2) unreliable video pseudolabels for model training. To address these limitations, we present a novel self-paced iterative learning model to enhance clustering and localization training simultaneously, thereby facilitating more effective unsupervised TAL. Concretely, we improve the clustering confidence through exploring the contextual feature-robust visual information. Thereafter, we design two (constant- and variable- speed)  incremental instance learning strategies for easy-to-hard model training, thus ensuring the reliability of these video pseudolabels and further improving overall localization performance. Extensive experiments on two public datasets have substantiated the superiority of our model over several state-of-the-art competitors.
\end{abstract}




\maketitle

\section{Introduction}
\begin{figure}[t!]
	\centering
	\subfigure[An example of TAL on a real-world surveillance video]{
		\label{fig:introduction:a} 
		\includegraphics[width=0.85\columnwidth]{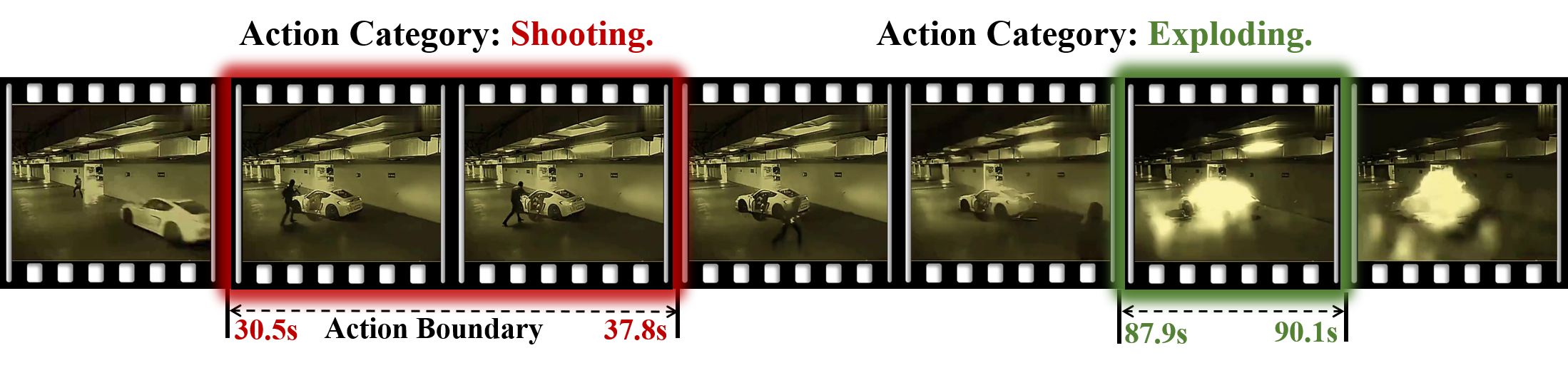}}
	\vspace{-5pt}
	\centering
	\subfigure[Three types of TAL model training under different supervisions]{
		\label{fig:introduction:b} 
		\includegraphics[width=0.85\columnwidth]{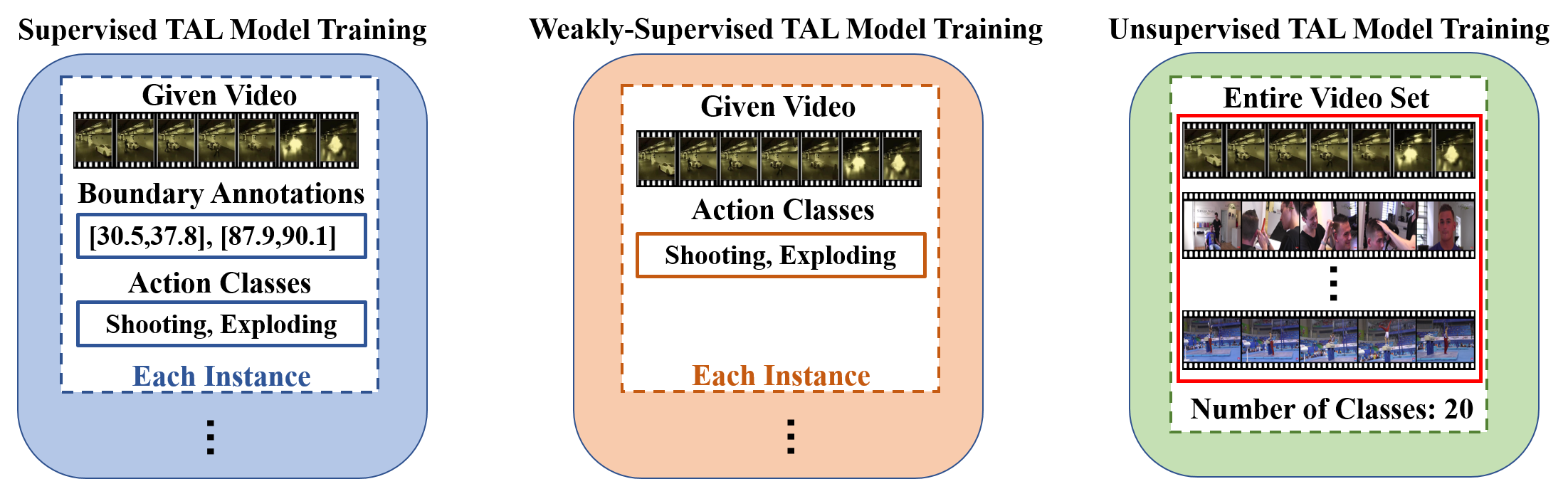}
	}
	\caption{Illustration of the temporal action localization task. }
	\vspace{-5pt}
	\label{fig:intro}
\end{figure}

With the rapid growth of videos in social media, video retrieval is always a hot yet challenging research topic over the past decades in the information retrieval \cite{zhang2025multi, han2022adversarial, shen2019video, yin2024exploiting, liu2023advancing, Liu2021IMP_GCN, chen2019relation,zhou2020temporal, sun2021exploiting,9404867, tang2021frame}. Traditional video retrieval identifies the most relevant video from a large collection of video candidates via a given query. Considering the diversity of visual content contained in the given video, users may be more concerned about a clip with specific action behaviors. As illustrated
in Fig~\ref{fig:introduction:a}, to secure the criminal evidence, police officers may pay close attention to the action ``Shooting'' and ``Exploding''. Therefore, localizing the temporal boundaries of target actions and identifying their categories within the given untrimmed video, i.e., temporal action localization~(TAL) \cite{shi2020weakly,zhang2021cola}, is highly desired in real-world application scenarios \cite{tang2021frame,hu2021coarse, li2024prototypical, wei2019neural, zhang2024somelvlm}.

Previous methods mainly rely on fully supervised TAL. As shown in Fig~\ref{fig:introduction:b}, to complete model training, these methods suffer from time-consuming and error-prone temporal action boundary annotation. Moreover, relatively subjective annotation results can also impede the overall localization performance. Although some  methods~\cite{shi2020weakly,lee2020background} are devoted to weakly supervised learning settings to reduce boundary annotation costs, they still require corresponding action category annotations, which are also labor-intensive.

Considering the above-mentioned defects, recent efforts have been dedicated to unsupervised temporal action localization (UTAL) \cite{yang2022uncertainty}, accomplishing the task of TAL only depending on the action class number of the entire training video set for model training, i.e., twenty actions in Fig~\ref{fig:introduction:b}. Specifically, Gong et al. and Yang et al. proposed the temporal class activation map (TCAM) model~\cite{gong2020learning} and the uncertainty guided collaborative training (UGCT) model~\cite{yang2022uncertainty} for UTAL, respectively. They uniformly adopted two-stage iterative `` clustering and localization'' settings, i.e., generating video-level pseudolabels and then training the localization model. Compared to supervised/weakly supervised TAL methods with high labeling costs, UTAL approaches offer greater scalability to cope with the continuous booming of videos. 

Despite its significance and value, UTAL is non-trivial due to the following two challenges: 1) \textbf{Clustering Confidence Improvement}. Both TCAM and UGCT need to conduct video clustering based pseudolabel generation. However, the former relying only on Euclidean distance based similarity measurement, is unable to ensure the correctness of generated pseudolabels. The latter may suffer from additional computational overhead from the mutual learning mechanism, and cannot guarantee the desired clustering effectiveness, especially in the case of semantic inconsistency between the extracted dual visual features. Therefore, how to improve clustering confidence is of vital importance for UTAL. 2) \textbf{Localization Training Enhancement}. The existing UTAL models directly use the full instance based iterative localization training. Considering the limitation of unreliable pseudolabeling, the full instance training strategy may bring in noise information and hurt the current localization training and the next clustering, thus causing superimposed harm on the overall ``iteratively clustering and localization''. Consequently, how to minimize the adverse effects of unreliable pseudolabels during iterative training is still a largely unsolved problem for UTAL.

To tackle these challenges, we propose a novel sel\textbf{F}-pac\textbf{E}d it\textbf{E}rative \textbf{L}earning model, dubbed as \textbf{FEEL}, for UTAL. {Inspired by the action category assumption~\cite{zhong2017re}, i.e., if the k-reciprocal nearest neighbors of two videos largely overlap, they likely contain the same action categories, we first present a Clustering Confidence Improvement (CCI) module to enhance clustering accuracy. Concretely, we introduce the feature-robust Jaccard distance measurement to estimate the semantic similarity based on the combined proximity of videos and their nearest neighbors, thus improving the clustering performance and assigning the true-positive instances to be top-ranked within each cluster. We then design an Incremental Instance Selection~{(IIS)} module for easy-to-hard iterative model training. Specifically, instead of directly employing the full instances, our module has the following two advantages: 1) it selects the most reliable video instances for model training during each iteration~(especially for the selection of the top-ranked ones by CCI module during the initial iterations), thereby minimizing the negative effects of unreliable pseudolabels; and 2) our module adopts constant and variable speed incremental selection strategies to adaptively select corresponding instances for targeted iterative model training, thus ensuring continuous performance improvement.} Finally, the target action can be effectively localized through adequate self-paced incremental instance learning. The main contributions of the FEEL method are as follows:
\begin{itemize}
	\item We present a novel self-paced iterative learning approach for UTAL. It selects the most reliable instances via constant and variable speed selection rates for iterative localization training, thus improving overall localization performance. To the best of our knowledge, it is the first attempt on integrating self-paced learning into UTAL. 
	
	\item  We introduce a feature-robust clustering confidence improvement module to enhance the clustering process, which synergizes with IIS module to bolster the generation of high-quality pseudolabels.
 

	\item We perform extensive comparison experiments, ablation studies, hyperparameter analysis, and visualizations to validate the promising performance of our model. We have released the involved codes and data to facilitate other researchers\footnote{Our codes and data: \url{https://github.com/tanghaoyu258/FEEL}.}.
\end{itemize}

\section{Related Work}


\subsection{Weakly-supervised TAL}
Weakly-Supervised TAL (WS-TAL) has been gaining popularity recently, since only the video-level action labels are needed for model training. The existing WS-TAL approaches mainly follow a general process: 1) generate the class-activation attention map from all snippets in the given video with a neural network, and 2) achieve the action classification and localization by thresholding on the attention map. Those WS-TAL approaches predominantly rely on either multiple instance learning \cite{hong2021cross, paul2018w} or temporal attention modeling \cite{zhang2021cola, nguyen2019weakly, wang2017untrimmednets} strategies. Specifically, on the one hand, the multiple instance learning methods aim to enhance intra-class feature representations by employing various losses \cite{liu2019completeness}. The temporal attention modeling methods, on the other hand, employ an attention mechanism to distinguish between action and non-action snippets in the video \cite{lee2020background,shi2020weakly, zhai2020two}. 

Although these existing WS-TAL approaches have achieved inspiring progress, they necessitate the annotations for video-level action categories of each video, i.e. their labeling cost is still high. In light of this, our proposed FEEL model can adaptively generate the corresponding action pseudolabels of the given videos, thereby seamlessly integrating with existing WS-TAL methods (e.g., CoLA\cite{zhang2021cola}) for effective UTAL.


\subsection{Unsupervised TAL}
Considering the limitations of WS-TAL, some efforts have been dedicated to unsupervised temporal action localization (UTAL). Gong et al. introduced the first UTAL {model, i.e., TCAM} \cite{gong2020learning}. {It first aggregates all training videos for video-level pseudolabel generation, and then adopts a temporally co-attention model with action-background separation loss and clustering-based triplet loss for action localization}. Following the similar settings, Yang et al. \cite{yang2022uncertainty} proposed a UGCT model to generate the pseudo label by collaboratively promoting the RGB and optical flow features, and then reduce the noise of pseudolabels through uncertainty awareness. In summary, these two models uniformly adopt the ``iteratively clustering and optimizing'' mechanism, i.e., iteratively generating the corresponding pseudolabels through the Euclidean-distance based clustering, and then training the localization model with all labeled instances.  According to the above analysis, the existing UTAL models still have shortcomings in clustering confidence and localization training. 

To overcome these impediments, our proposed FEEL model utilizes CCI to refine pseudolabel generation, and dynamically selects the most reliable labeled videos instead of the entire instances for localization training, both of which enable the superiority of FEEL over the existing methods.

\subsection{Self-paced learning paradigm}
Inspired by the human learning process, i.e., knowledge can be acquired through easy-to-difficult curriculum learning, Bengio et al. presented the Curriculum Learning~(CL) paradigm where knowledge is learned step by step in an easy-to-difficult manner, under the guidance of a pre-defined criterion. Since CL theory requires a prior indicator to determine the hardness of an instance, the self-paced learning (SPL)~\cite{kumar2010self} is investigated to incorporate the automatic hardness determination into the model training. Theoretical analysis has proven that the SPL paradigm is capable of preventing the latent variable model from falling into the bad local optimums or oscillations~\cite{ma2017self}. Due to its effectiveness, the SPL paradigm, under semi-supervised or unsupervised settings, has been used in some research topics like image classification \cite{caron2018deep} and person re-id \cite{fan2018unsupervised,wu2019progressive}. For example, Fan et al. \cite{fan2018unsupervised} proposed to cluster the features of pedestrians and train the Resnet \cite{he2016deep} extractor with the generated pseudo labels iteratively. Caron et al. \cite{caron2018deep} introduced a DeepCluster network that integrates the image feature clustering to the optimization of parameters in the convolutional network.


To the best of our knowledge, our proposed FEEL is the first attempt to address UTAL with the self-paced learning paradigm. Considering that unreliable pseudolabeling may lead to local optimum during model training, our model iteratively selects the most reliable instances for localization training based on constant-speed and variable-speed, respectively. Moreover, we propose to refine the label predictions through the feature-robust distance measurement, which fits the SPL process well.

\begin{figure*}[t]
	\centering
	\includegraphics[width=0.88\textwidth]{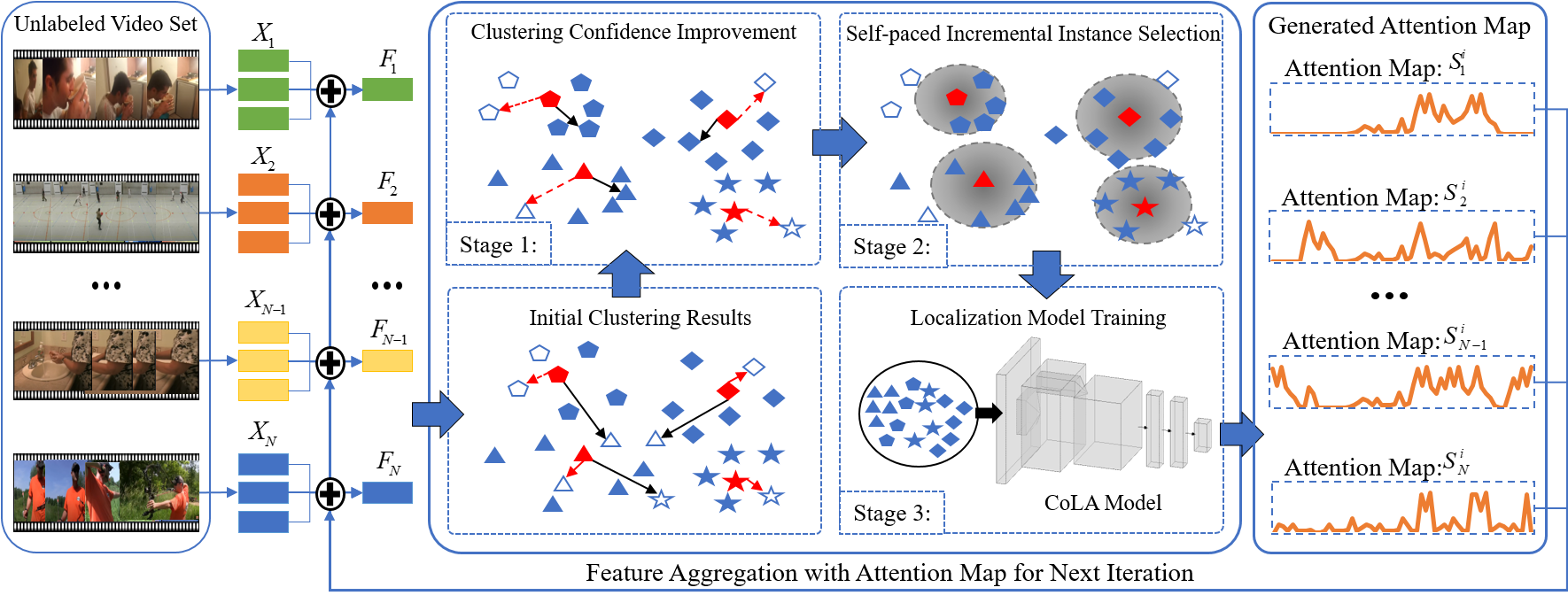}
	\caption{{An illustration of our FEEL model. Based on the initial clustering results, it conducts three stages within each iteration: adopting the CCI to refine the initial clustering for pseudolabel generation; employing IIS to select the most reliable instances for localization training; localization model training.} {Within the iteration, we employ distinct shapes to distinguish different clusters. Besides, a solid dot means that the corresponding video is correctly pseudolabeled, while a hollow dot means the opposite. The red, solid dots specifically denote the clustering centers of each cluster. As we can see, the CCI module corrects some mislabeled videos, and simultaneously pulls correctly labeled instances closer to the clustering centers while moving erroneously labeled ones farther away. Afterward, only the videos with high-labeling quality (the dots within the shaded region) are selected for model training.}} 
	\label{fig:model}
\end{figure*}

\section{Our Proposed FEEL Model}  
\subsection{Preliminary}
In this section, the necessary denotations are detailed for UTAL task. Given the training video set $\mathcal{V}=\{v_n\}^{N}_{n}$ from $K$ action classes, we divide each video $v_n$ into a fixed number of 16-frame non-overlapping video snippet $S_n=\{s_{n,t}\}_{t=1}^{T}$, where $N$ denotes the number of untrimmed videos, and $T$ denotes the number of video snippets. {Following the previous practices \cite{zhang2021cola,nguyen2018weakly}}, we adopt the pre-trained feature extraction network to separately embed $S_n$ into the RGB features ${X}^{R}_{n}=\{x^{R}_{n,t}\}_{t=1}^{T} \in R^{T*d}$ and optical flow features ${X}^{O}_{n}=\{x^{O}_{n,t}\}_{t=1}^{T} \in R^{T*d}$. Subsequently, we concatenate ${X}^{R}_{n}$ and ${X}^{O}_{n}$ along the temporal dimension to formulate the final video snippet representations  ${X}_{n}=\{x_{n,t}\}_{t=1}^{T} \in R^{T*2d}$, where $x_{n,t}$ denotes the $t$-th snippet feature and $d$ denotes the feature dimension. Under the unsupervised settings, since video-level groundtruth action labels are unavailable, each video $v_n$ in the training set $\mathcal{V}$ is assigned a generated video-level pseudolabel $\tilde{y}_n$, so that the unsupervised setting is converted to the weakly-supervised one, and a general weakly-supervised localization model $\mathcal{M}$ can be trained for localization based on the pseudolabeled videos.

As depicted in Fig \ref{fig:model}, our FEEL method operates this unsupervised process in an iterative manner. During the $i$-th iteration, our method is initialized by generating the initial clustering results for the global video features $\{	\mathbf{F}_{n} \}^{N}_{n=1}$ in the training set $\mathcal{V}$. Particularly, for the $n$-th video $v_n$,  given the varying importances of the snippets $\{x_{n,t}\}_{t=1}^{T}$ in ${X}_{n}$, the global video feature $\mathbf{F}_{n}$ is computed by adaptively summarizing the snippets in ${X}_{n}$, using the corresponding class-agnostic attention map $\mathbf{S}^{i-1}_{n} \in R^{T}$, as follows:
\begin{equation}
	\mathbf{F}_{n} =\sum\nolimits_{t=1}^{T}({s}^{i-1}_{n,t} \cdot {x}_{n,t}) 
\end{equation}
where the iteration stage $i$ is omitted in $\mathbf{F}_{n}$ and following notations for simplicity, and $\mathbf{S}^{i-1}_{n} \in R^{T}$ is produced through the weakly-supervised model $\mathcal{M}$ from $(i-1)$-th iteration. We detailed the structure of $\mathcal{M}$ in the section \ref{section:method_WSTALmodel}.  After all global video features $\{	\mathbf{F}_{n} \}^{N}_{n=1}$ are obtained, we perform a clustering algorithm based on Euclidean distance (e.g., K-means) that divides those global video features ${ \mathbf{F}_{n}}$ into $K$ clusters, each with a cluster center $c_{k}$ representing a pseudo action class ${y}_{k}$. For $c_{k}$ in the cluster center set $\mathcal{C}=\{c_k\}^{K}_{k=1}$, its Euclidean distance ${d}_{E}(c_k,{v}_{n})$ to each video ${v}_{n}$ in the video set $\mathcal{V}$ is calculated. Thereafter, a confidence matrix $\mathcal{D}_{E} \in R^{K*N}$ is constructed with the paired distances, where each entry is denoted as the initial pseudolabeling confidence of ${v}_{n}$ to the class $c_{k}$.

Based on the initial clustering results, our FEEL model proceeds in the following three stages, shown in Fig \ref{fig:model}. 1) Employing the {clustering confidence improvement module to refine the initial pseudolabels}. 2) Incrementally selecting the most reliable video instances ( the dots within the shaded region in the figure) according to our constant- and variable-speed selection criterion. 3) Training a localization model based on these selected pseudolabeled instances and generating the class-agnostic attention map for pseudolabel prediction in the next iteration. The details of these three stages are elaborated upon sequentially in the following sections.

\subsection{Clustering Confidence Improvement}
\label{section:method_CCI}
Under the UTAL constraint, the video-level annotations are unavailable. Therefore, it is necessary to produce the pseudo action label for each video. Based on the clustering results of the Euclidean distance ${d}_{E}(c_k,{v}_{n})$ calculated from the global video features $\mathbf{F}_{n}$, the existing methods directly assign the pseudo action label ${y}_k$ by the nearest clustering centroid $c_k$ for each video ${v}_{n}$ and select all labeled videos for model training during each iteration. This operation raised two major problems: 1) The unsatisfactory video-level annotations. The global video feature $\mathbf{F}_{n}$ of each video is obtained through the feature attentive aggregation across all its snippets, which is often unsatisfied due to inferior generated attention map at the early iterations, so the reliability of the Euclidean distance calculated between the global features cannot be guaranteed. 2) Many mislabeled videos will be top-ranked in each cluster. Due to the inaccuracy of Euclidean distance labeling confidence, a large number of mislabeled videos in a cluster will be closer to its center, i.e., have higher clustering confidence. Under this condition, despite that our selection strategy can dynamically increase the number of selected high confidence videos as the iteration goes, these top-ranked mislabeled videos will be selected in the early iteration, which significantly pollutes the early model training. To address the above issues, we introduce a Clustering Confidence Improvement module to achieve high-quality video labeling. 

Given the initial pseudo labeling confidence matrix $\mathcal{D}_{E} \in R^{K*N}$, we rank each row in $\mathcal{D}_{E}$ ascendingly. The objective is to enhance the obtained initial sorting list $\mathcal{R}_{k} =[{v}_{1},{v}_{2},...,{v}_{N}]$ for each cluster center $c_{k}$, so that more true-positive videos will be top-ranked than false-positive data in each list. Thereafter, the labeling accuracy of all training videos, especially of those top-ranked ones, will be refined, and thus the model optimization will be improved. The $l$-reciprocal nearest neighbors \cite{zhong2017re} have proven effective in achieving this objective by capturing the contextual cues among the distribution of video features in the feature space. Formally, we first formulate the $l$-reciprocal nearest neighbors of $c_{k}$ as:
\begin{equation}
	\mathcal{U}(c_{k}, l)=\{{v}_{n} \mid({v}_{n} \in \mathcal{N}(c_{k}, l)) \wedge(c_{k} \in \mathcal{N}({v}_{n}, l))\}
\end{equation}
where $\mathcal{N}(c_{k},l)$ represents the top-$l$ neighbors of $c_{k}$ in the initial list $\mathcal{R}_{k}$. Compared with the initial list $\mathcal{R}_{k}$, the $l$-reciprocal nearest neighbors $\mathcal{U}(c_{k}, l)$ requires both $c_k$ and ${v}_{n}$ to be $l$-nearest neighbors of each other, which ensures true-matches between them to a greater extent. Since this rule is too strict that some hard positive samples will also be filtered out, the  $\frac{1}{2}l$-reciprocal nearest neighbors of each sample in  $\mathcal{U}(c_{k}, l)$ is included to form a new neighboring set $\hat{\mathcal{U}}(c_{k}, l)$ as:
\begin{equation}
	\begin{aligned}
		& \hat{\mathcal{U}}(c_{k}, l) = \mathcal{U}(c_{k}, l) \cup \mathcal{U}(z, \frac{1}{2}l)\\
		& 	\text { s.t. } |\mathcal{U}(c_{k}, l) \cap \mathcal{U}(z, \frac{1}{2}l)| \geq \frac{2}{3} |\mathcal{U}(z, \frac{1}{2}l)|\\
		& \forall~~z \in \mathcal{U}(c_{k}, l)
	\end{aligned}
\end{equation}
Compare to ${\mathcal{U}}(c_{k}, l)$,  the incremented set $\hat{\mathcal{U}}(c_{k}, l)$ takes more positive videos into account. Intuitively, a video is more likely to be matched with a clustering center only when there are more common samples in their $l$-reciprocal nearest neighbor sets. Following this principle, we introduce the feature-robust Jaccard distance \cite{zhong2017re} that measures the interaction over union between the $l$-reciprocal sets of $c_{k}$ and ${v}_{n}$ as follows: 
\begin{equation}
	\label{equation:jaccard_set}
	d_{J}(c_{k}, {v}_{n})=1-\frac{|\hat{\mathcal{U}}(c_{k}, l)  \cap \hat{\mathcal{U}}({v}_{n}, l)|}{|\hat{\mathcal{U}}(c_{k}, l)  \cup \hat{\mathcal{U}}({v}_{n}, l)|}
\end{equation}
where $|\cdot|$ represents the size of the set. Since calculating the interaction over union of $\hat{\mathcal{U}}(c_{k}, l)$ and $\hat{\mathcal{U}}({v}_{n}, l)$ for all center and unlabeled video pairs is exhaustive, we encode the l-reciprocal nearest neighbor set $\hat{\mathcal{U}}(c_{k}, l)$ into an embedding $\mathbf{E}_{c_{k}}=[e_{{c_k};{{v}_{1}}},e_{{c_k};{{v}_{2}}},...,e_{{c_k};{{v}_{N}}}]$, so that the overlapping calculation between two sets can be transferred to the vector operation. Besides, considering that the importance of neighbors in different positions should be discriminated against, we assign the nearer neighbors a larger value than that of the farther ones in the embedding $\mathbf{E}_{c_{k}}$. Formally, the embedding $\mathbf{E}_{c_{k}}$ is defined as:
\begin{equation}
	e_{{c_k};{{v}_{n}}}=\left\{\begin{array}{ll}
		\mathrm{exp}\left({-{d}_{E}\left(c_k,{v}_{n}\right)}\right), & \text { if } ~~{v}_{n} \in \hat{\mathcal{U}}(c_{k}, l) \\
		0, & \text { otherwise }
	\end{array}\right.
\end{equation}
Similarly, the $l$-reciprocal nearest neighbor set $\hat{\mathcal{U}}({v}_{n}, l)$ of ${v}_{n}$ is transferred to the embedding $\mathbf{E}_{{v}_{n}}$, and the calculation of the Jaccard distance can be represented as:
\begin{equation}
	d_{J}(c_{k}, {v}_{n})=1-\frac{\sum_{i=1}^{N} \min (e_{{c_k};{{v}_{i}}}, e_{{{v}_{n}};{{v}_{i}}})}{\sum_{i=1}^{N} \max (e_{{c_k};{{v}_{i}}}, e_{{{v}_{n}};{{v}_{i}}})}
\end{equation}
where $\min(\cdot,\cdot)$ and $\max(\cdot,\cdot)$ identify the minimum and maximum value in the set, respectively. To better measure the similarity relationship between the two videos, the Jaccard distance is integrated into the original Euclidean distance for the final refined distance as the labeling criterion, which can be formulated as:
\begin{equation}
	\label{equation:CCI_loss}
	d(c_{k}, {v}_{n})= \gamma d_{J}(c_{k}, {v}_{n}) + (1-\gamma){d}_{E}(c_{k}, {v}_{n})
\end{equation}
where $\gamma$ controls the contributions of the original Euclidean distance ${d}_{E}$ and the Jaccard distance $d_{J}$.  After all pairwise refined distance between the cluster center $c_{k}$ in $\mathcal{C}$ and the unlabeled video ${v}_{n}$ in $\mathcal{V}$ is computed, we rerank the initial list $\mathcal{R}_{k}$ of each clustering center as $\hat{\mathcal{R}}_{k} =[\hat{v}_{1},\hat{v}_{2},...,\hat{v}_{N}]$, which is adopted for Self-paced Incremental selection in next section. 


\subsection{Self-paced Incremental Instance Selection}
In this section, a self-paced incremental selection strategy is introduced to identify the most reliable pseudolabeled videos. In the previous methods, the entire training set labeled by clustering is selected for further localization model training. Since the clustering results severely depend on the attention map generated from the weak localization model, the video annotations are inaccurate in the early iterations, especially for those challenging videos. Although our CCI module has significantly improved the quality of all video annotations, the rest mislabeled videos will still limit the learning process, leading the model into a bad local optimum. Therefore, we introduce a self-paced incremental selection strategy, which progressively samples an increasing number of labeled videos from easy to hard as the localization model becomes robust during the iterations. Particularly, at the $i$-th iteration, regarding the refined distance $d(c_{k}, {v}_{n})$ as labeling criterion, we assign the pseudo action label for unlabeled video ${v}_{n}$ by its nearest clustering center, which is formulated as:
\begin{equation}
	\hat{c}_{k}, \hat{y}_{k}=\arg \min _{\left(c_{k}, y_{k}\right) \in \mathcal{C}}d(c_{k}, {v}_{n})
\end{equation}
\begin{equation}
	\label{equation:label_assign}
	\tilde{y}_n = \hat{y}_{k}
\end{equation}
where $\hat{c}_{k}$ and $\hat{y}_{k}$ denote the nearest center to ${v}_{n}$ and the corresponding pseudo action class, respectively. Based on the obtained tuple of labeled video $({v}_{n},\tilde{y}_n)$, we filter out all videos with different action labels from the reranking list  $\hat{\mathcal{R}}_{k}$ to generate the new ranking list ${\mathcal{R}}^{*}_{k}=[{v}^{*}_{1},{v}^{*}_{2},...,{v}^{*}_{n_{k}}]$ of each center $c_{k}$, where $n_k$ denotes the number of remained videos in $k$-th cluster. In this way, each video ${v}_{n}$ will appear in only one of the ranking lists ${\mathcal{R}}^{*}_{k}$ of those $K$ centers. At the $i$-th iteration, we sample several top-ranked videos of each center ${c}_{k}$ into the selected pseudo-labeled video set ${V}^{*}_i$ as follows:
\begin{equation}
	{V}^{*}_{i,k}=\{{v}^{*}_{j}\mid {v}^{*}_{j} \in {\mathcal{R}}^{*}_{k},  1 \leq j \leq \beta_{i,k}\cdot n_{k}\}
\end{equation}
\begin{equation}
	{V}^{*}_{i}={V}^{*}_{i,1}\cup {V}^{*}_{i,2}\cup\cdot\cdot\cdot\cup{V}^{*}_{i,K-1}\cup{V}^{*}_{i,K}
\end{equation}
where $\beta_{i,k} \in [0,1]$ denotes the percentage of selected videos from $k$-th action at the $i$-th iteration. According to our incremental selection strategy, at the initial iteration, $\beta_{i,k}$ is relatively small so that a small fraction of the top-ranked videos are selected to enable the reliability of the selected videos. As the pseudo labels become more reliable during the iterations, more hard and diverse videos are included, and $\beta_{i,k}$ gradually grows to 1 with the selected set ${V}^{*}_{i}$ enlarged to the entire training video set $\mathcal{V}$. Obviously, it is crucial to control the enlarge speed of $\beta_{i,k}$, so that the quality of the selected samples is ensured and enough samples for the training of the weakly supervised model are retained as well. Thereafter, we set the same selection rate for all actions for simplicity, i.e., $\beta_i = \beta_{i,1}=\beta_{i,2}=\cdot\cdot\cdot=\beta_{i,K}$, and introduce two different incremental selection strategy in this paper: (1) constant mode: the selection rate $\beta$ is increased linearly during the iteration, i.e., $\beta_i= i/I_{max}$, where $I_{max}$ denotes the total iteration number;
(2) variable mode: the selection rate $\beta_i$ is increased following a concave curve function, which is expressed as:
\begin{equation}
	\beta_i=\frac{\mu^{i}-1}{{\mu^{I_{max}}}-1}
\end{equation}
where $\mu$ controls the concavity of this curve. 
\begin{figure}[!t]
	\centering
	\includegraphics[width=0.85\columnwidth]{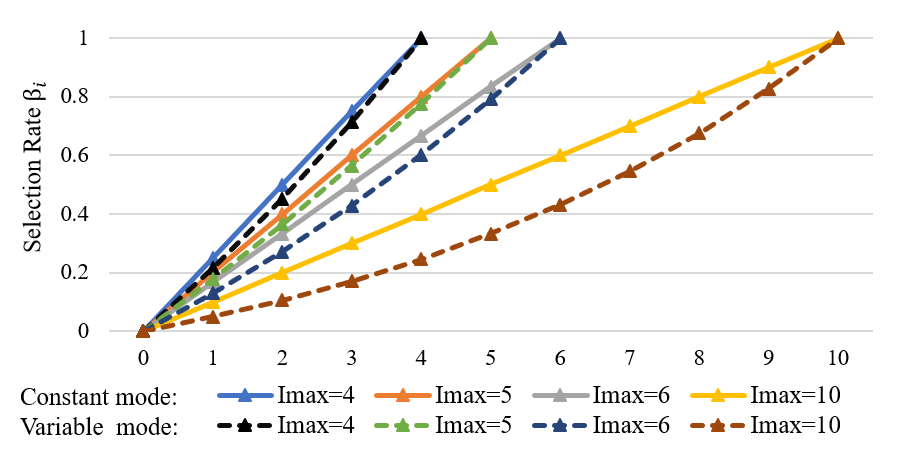}
	\caption{Illustrations of the enlarging curves with different $I_{max}$, where the solid and dashed lines are the constant mode and variable mode, respectively.}
	\label{fig:model_selectionmode}
\end{figure}


\textbf{Discussion}: Fig \ref{fig:model_selectionmode} illustrates the enlarging modes of those two strategies with different $I_{max}$, where the horizontal axis represents the iterations and the vertical axis signifies the proportion of selected video relative to the entire training set. For the constant mode, the size of the selected video set $|{V}^{*}_{i}|$ increases at the same speed, which is controlled by the total iterations $I_{max}$. If we choose a large $I_{max}$, the selected video set $|{V}^{*}_{i}|$ enlarges only by a small fraction of videos per iteration, which indicates a more stable data quality and growth of localization performance. Besides, a relatively small $I_{max}$ indicates the number of selected videos $|{V}^{*}_{i}|$ grows rapidly, resulting in a faster training process and a decrease in annotation accuracy of the selected videos for model training.

Compared to the constant mode which maintains the same enlarging speed during the entire process, the variable mode grows slower at the initial iterations and then faster. When we set the same $I_{max}$ for both modes, on the one hand, the variable mode enables higher reliability of the selected samples in the initial iteration. As the localization model becomes robust at the later iterations, the faster growth rate will reduce the overall training time. {On the other hand, it is crucial to control the concavity of this mode. If the concavity is set too large, the initially selected training samples will be too few to sufficiently train the existing weakly supervised models that are often trained based on contrastive loss, which will affect the localization performance in the later iterations. In fact, the experiments demonstrate that the constant mode can already achieve a promoting performance while the variable mode performs even better if the concavity is controlled carefully. }

\begin{figure}[t]
	\centering
	\includegraphics[width=0.95\linewidth]{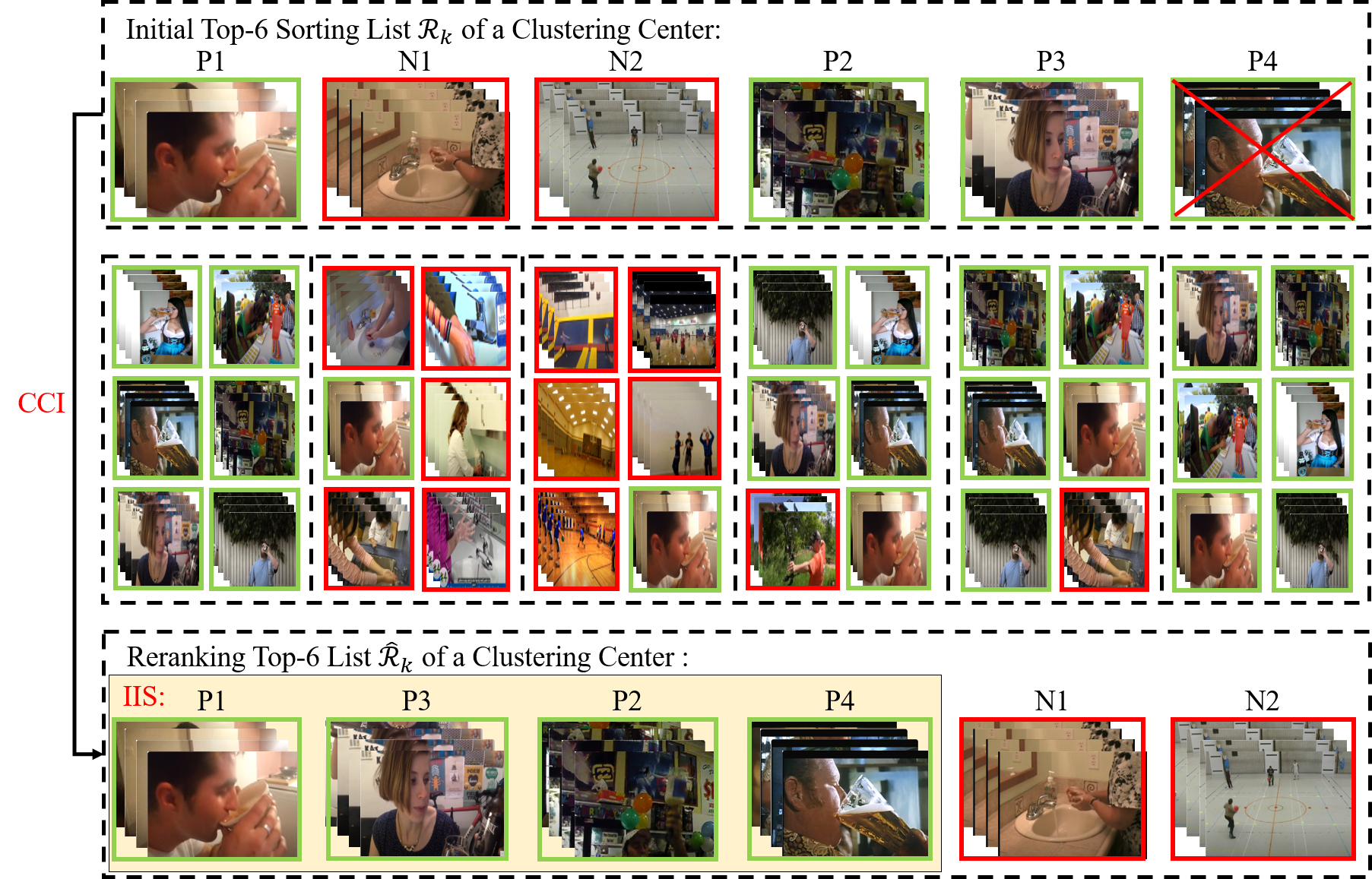}
	\caption{Illustration of CCI and IIS module on an action cluster, where the positive videos of this cluster are marked in green rectangle. \textbf{Top}: The initial top-6  ranking list of a clustering center, where P1-P4 are positives, N1-N2 in red rectangle are negatives. P4 marked with $\times$ means this positive video is falsely labeled to other action. \textbf{Middle}: Each two columns represents the top-6 neighbors of the corresponding video. It is evident that a significant overlap exists between the top-6 neighbors of P1-P4 and those of the clustering center. \textbf{Bottom}: The reranking top-6 list of this cluster. Based on IIS module, only the top-4 videos of this list, which are highlighted in yellow, are selected for model training.}
	\label{fig:modelidea}
\end{figure}

\textbf{Discussion of the synergistic effects of CCI and IIS}: Fig \ref{fig:modelidea} illustrates the CCI and IIS module operating on an action cluster during an iteration, where the positive videos are marked in green and negative ones are marked in red. As we can see, according to the existing methods, the negative video N1 and N2 are ranked higher than P2-P4, and P4 is falsely labeled to another action in the initial sorting list $\mathcal{R}_{k}$. However, given the nearest neighbors of all videos, our CCI refines $\mathcal{R}_{k}$ so that P1-P4 are ranked higher than the negatives in the reranking list $\hat{\mathcal{R}}_{k}$, and the pseudolabel of P4 is also corrected. Thereafter, our IIS module dynamically selects the top-ranked four videos (highlighted in yellow) out of six, which is 100\% correctly labeled.  The combined effects of CCI and IIS contribute significantly to the high-quality training of the localization model.

\subsection{Temporal Action Localization Training}
\label{section:method_WSTALmodel}
With the most reliable labeled videos selected in ${V}^{*}$, the localization model can be easily trained end-to-end. Note that our method focuses on improving the quality of the selected pseudo-labeled samples for the unsupervised TAL task, and thus does not depend on any specific attention-based localization model. Here we directly adopt the existing Contrastive learning to Localize Actions (CoLA) \cite{zhang2021cola} model $\mathcal{M}$ . More formally, at the $i$-th iteration, this model first embeds the snippet features ${X}_{n}=\{x_{n,t}\}_{t=1}^{T} \in R^{T*2d}$ of the video ${v}^{*}_n$ in the selected video set ${V}^{*}={({v}^{*}_n,\hat{y}_{k})}$ into ${X}^{E}_{n}=\{x^{E}_{n,t}\}_{t=1}^{T} \in R^{T*2d}$ through a linear layer followed by the ReLU function. Taking ${X}^{E}_{n}$ as input, we compute a series of the class-specific attention map ${A}^{i}_{n}$ as follows:
\begin{equation}
	{A}^{i}_{n} = \delta(\mathbf{W}_{c}{X}^{E}_{n}+\mathbf{b}_{c}) 
\end{equation}
where $\mathbf{W}_{c}$ and $\mathbf{b}_{c}$ are learnable parameters, and $\delta$ denotes the ReLU activation function. The $k$-th column of ${A}^{i}_{n} \in R^{T*K}$ represents the probability of action class $k$ occurring along the temporal dimension. To model the actionness of each snippet, we adopt the column-wise addition and a followed Sigmoid activation function, which is expressed as ${S}_{n}^{i} = Sigmoid(f_{add}({A}^{i}_{n}))$, to obtain the class-agnostic attention map ${S}_{n}^{i} \in {R}^{T}$. The obtained attention map ${S}_{n}^{i} \in {R}^{T}$ is then adopted for global feature aggregation of video $v_{n}$ in the next iteration. Note that ${S}_{n}^{i}$ is inaccessible at the first iteration, we simply define ${S}_{n,t}^{1} = 1/T ( 1 \leq t \leq T)$.   

Based on the attention map ${S}_{n}^{i} \in {R}^{T}$, the localization model mines hard video snippets and easy video snippets for contrastive learning. Specifically, on the binary sequence generated by setting a threshold $\tau_c$ on ${S}_{n}^{i}$, the expansion or erosion operations \cite{zhang2021cola} are performed to expand or reduce the interval range of boundary adjacent action proposals, and the hard action snippets ${X}^{hf}_{n} \in R^{{T}^{hard}*2d}$ and hard background snippets ${X}^{hb}_{n} \in R^{{T}^{hard}*2d}$ are obtained. Besides, the snippets with the top-${T}^{easy}$ and bottom-${T}^{easy}$ attention scores are regarded as the easy action snippets ${X}^{ef}_{n} \in R^{{T}^{easy}*2d}$ and easy background snippets ${X}^{eb}_{n} \in R^{{T}^{easy}*2d}$, respectively. More details of mining the action and background snippets can be found in \cite{zhang2021cola}.

With the selected pseudo labeled video set ${V}^{*}$ and the mined easy and hard video snippets, the CoLA model is trained end to end with the objective function comprised of two parts:
\begin{equation}
	\label{equation:WTAL_loss}
	{L}={L}_{cls}+\lambda{L}_{ctr}
\end{equation}
where $\lambda$ balances those two losses. Firstly, ${L}_{cls}$ represents the commonly used classification loss. As in \cite{wang2017untrimmednets,lee2020background,zhang2021cola}, the top-$l^{high}$ largest attention scores in each raw of ${A}^{i}_{n}$ are averaged to get  the video-level action prediction $a^{i}_{n} \in {R}^{K}$ for all classes. Regarding pseudo label $\tilde{y}_n$ in Eq.~\ref{equation:label_assign} as the ground-truth action, the action prediction $a^{i}_{n}$ is fed into a softmax function to obtain the action class probabilities $p^{i}_{n}$, and this loss maximizes the $k$-th class probabilities $p^{i}_{n;k}$ as:
\begin{equation}
	{L}_{cls}= -\frac{1}{N} \sum_{n=1}^{N} \sum_{k=1}^{K} \tilde{y}_n\log(p^{i}_{n;k})
\end{equation}
The second loss term ${L}_{ctr}$ is the snippet-level contrastive learning loss that refines the representations of hard snippets. More formally, the hard action and hard background pairs for contrastive learning are separately formed. For the hard action pair, we sample a query feature ${X}^{hf}_{n,t} \in {R}^{1*2d}$, a positive feature ${X}^{ef}_{n,t^{+}} \in {R}^{1*2d}$, and  $T^{-}$ negative features $\{{X}^{eb}_{n,t^{-}}\}^{T^{-}}\in {R}^{{T^{-}}*2d}$ from the mined hard action ${X}^{hf}_{n}$, easy action ${X}^{ef}_{n}$, and easy background ${X}^{eb}_{n}$, respectively. As for the hard background pair, the query feature, positive feature, and negative features are sampled from the hard background ${X}^{hb}_{n}$, easy background ${X}^{eb}_{n}$, and easy action ${X}^{ef}_{n}$, respectively. Thereafter, the contrastive loss can be constructed as:
\begin{equation}
        \begin{aligned}
		&\mathcal{L}\left({X}_{n},{X}_{n}^{+},{X}_{n}^{-}\right) \\
		& =-{\log \frac{\exp \left({X}_{n} \cdot {X}_{n}^{+} / \theta \right)}{\exp \left({X}_{n} \cdot {X}_{n}^{+} / \theta \right)+\sum^{T^{-}}_{{t^{-}}=1} \exp \left({X}_{n} \cdot {X}_{n}^{-} /  \theta \right)}}
	\end{aligned}
\end{equation}
\begin{equation}
        \begin{aligned}
		&{L}_{ctr}  =-\sum^{N}_{n=1}{\mathcal{L}\left({X}^{hf}_{n,t},{X}^{ef}_{n,t^{+}},{X}^{eb}_{n,t^{-}}\right) }\\
		& -\sum^{N}_{n=1}{\mathcal{L}\left({X}^{hb}_{n,t}, {X}^{eb}_{n,t^{+}},{X}^{ef}_{n,t^{-}}\right) }
	\end{aligned}
\end{equation}
where $\mathcal{L}\left({X}_{n},{X}_{n}^{+},{X}_{n}^{-}\right)$ is a defined distances computation process for three features ${X}_{n}$, ${X}_{n}^{+}$ and ${X}_{n}^{-}$. $\theta$ denotes the temperature factor that is defaulted to 0.07. Based on this loss term, the similarities between hard and easy action and between hard and easy background are maximized, which thereby refines the snippet feature representations.

\subsection{Inference}
After completing all the iterations, we use the trained localization model for inference. To facilitate the evaluation, it is necessary to map each cluster to the ground-truth action labels. Following previous methods~\cite{zeng2021end,ji2019invariant,park2021improving,gong2020learning}, we assign labels based on the predominant action category within each cluster. Note that this label assignment, while utilizing the labels of action category, does not constitute learning in the traditional sense. The unsupervised UTAL training process primarily relies on the numerical labels derived from the clustering to train the model. During the inference stage, the input video is first fed into the trained localization model to generate the class-specific attention map ${A}^{i}_{n}$ and the video-level action prediction $a^{i}_{n}$. By thresholding on $a^{i}_{n}$, we select all the action classes that satisfies $a^{i}_{n,k} \textgreater \tau$. For those selected actions, $\tau_a$ is adopted to threshold on the corresponding attention map ${A}^{i}_{n}$ to obtain the set of video proposal candidates. Since there might be many overlapping proposals across different actions, we apply the non-maximum suppression (NMS) with a threshold of 0.7 on those duplicated proposals for the final localization results. 

\subsection{Discussion on Training Time}
Despite the introduction of additional iterations in our FEEL, the overall training time is not substantially prolonged. This efficiency is achieved by the following aspects:

Adaptive Batches: To reduce the training time, our FEEL method takes the TCAM \cite{gong2020learning} as the example to adjust the number of training batches according to the proportion of selected pseudolabeled instances in each iteration, i.e., if only 10\% of the videos are selected within an iteration, our FEEL will adaptively reduce the number of training batches to 10\% of that of TCAM. This adaptive training batch strategy significantly mitigates the training time for each iteration.

Clustering Complexity: Our FEEL method employs the K-means algorithm and the CCI module for clustering, each with a complexity of $O(KN)$ and $O({N}^{2}\operatorname{log}{N})$, respectively. The combined complexity for clustering in our method is thus $O({N}^{2}\operatorname{log}{N})$, which is computationally more efficient than spectral clustering commonly leveraged by existing methods, characterized by a complexity of $O(N^3)$.

\textbf{Iteration Numbers}: It is noteworthy that while existing methods only report the localization results of the three iterations, we have observed that further iterations do not yield performance gains for these methods due to the poor pseudolabel quality. In contrast, FEEL demonstrates consistent performance improvements across more iterations.

In our practical implementations, the training time of FEEL, when set $I_max=6$, is approximately equivalent to that of TCAM \cite{gong2020learning}. At this juncture, our FEEL has successfully achieved notable performance across both datasets. This observation further verifies that our FEEL method manages to enhance performance without significantly extending the overall training time.

\section{Experiments}
In this section, we first introduce the experimental settings. And then, we perform comparison experiments, ablation studies, and  hyper-parameter analysis to answer the following 4 research questions (\textbf{RQs}) sequentially:

\begin{itemize}
	\item \textbf{RQ1:} Is our model FEEL able to exceed several state-of-the-art UTAL competitors?
	\item \textbf{RQ2:} Is each component of FEEL contributed to boost the localization performance?
	\item \textbf{RQ3:} How do the iteration process variant $I_{max}$ and $\mu$ affect the overall performance of our FEEL model?
	\item \textbf{RQ4:} {Is our FEEL well scalable on UTAL}?
\end{itemize}

\subsection{Experiment Setup}
\subsubsection{Dataset} 
\textbf{THUMOS'14} \cite{idrees2017thumos}: This dataset consists of 200 and 213 untrimmed videos for validation and testing, respectively, which includes 20 action classes in total. Each video contains an average of 15 action segments with temporal action boundary annotations. Following the conventional approach, we employ the validation data for model training and the test data for evaluation.

\textbf{ActivityNet v1.2} \cite{caba2015activitynet}: This large-scale video benchmark dataset collected for human activity understanding contains 4819, 2383, and 2480 videos for training, validating, and testing, respectively. Since the annotations of 2480 test videos are withheld, the 2383 videos in the validation set are treated as test data. In this dataset, each video has an average of 1.5 action segments labeled with temporal action boundaries, belonging to 100 different action classes.

\subsubsection{Implementation Details} 
For the videos, the pretrained I3D network \cite{carreira2017quo} adopted in CoLA \cite{zhang2021cola} is employed to extract the RGB and optical flow snippet features, both of which are with 1024-dimension. During the entire process, the parameters of the feature extractors are fixed. For the clustering confidence improvement, we set $l$ in $ \hat{\mathcal{U}}(c_{k}, l)$ and $\mathcal{N}({v}_{n}, l)$ to 20 and 6, respectively. $\gamma$ in Eq.\ref{equation:CCI_loss} is set to 0.7. $I_{max}$ is set among {4, 5, 6, 10} for both incremental selection mode, and $\mu$ of the variable mode is set to 1.05 and 1.03 for THUMOS'14 and ActivityNet v1.2, respectively. The hyperparameters in the localization model are set to the default parameters of the CoLA model \cite{zhang2021cola}. Specifically, the temporal length $T$ of each video is set to 50 and 750 for ActivityNet v1.2 and THUMOS'14  datasets, respectively. For the snippet contrastive learning, we set  $\tau_c = 0.5$, ${T}^{-}={T}^{easy} = \max (1,\lfloor{T/8}\rfloor)$, ${T}^{hard} = \max (1,\lfloor{T/32}\rfloor)$. $\lambda$ in Eq.\ref{equation:WTAL_loss} is set as 0.005. To make the training time as fair as possible, we proportionally set the number of training epochs $E_{i}$ in the $i$-th iteration of our method as: $E_{i} = E_{max}* {V}^{*}_{i}/\mathcal{V}$, where $E_{max}$ denotes the training epochs of other UTAL methods in a single iteration. During each iteration, the model is trained with a batch size of 128 and 16 for  ActivityNet v1.2 and THUMOS'14, respectively, and the Adam optimizer with a learning rate of 0.0001 is adopted.  During the inference stage, the action class threshold $\tau$ is set to 0.1 on ActivityNet v1.2 and 0.2 on THUMOS'14. $\tau_a$ is set to [0:0.15:0.015] and [0:0.25:0.025] for ActivityNet v1.2 and THUMOS'14 dataset, respectively.   

\subsubsection{Evaluation Metrics}
The standard evaluation metric mean Average Precision (mAP) under different interaction over union (IoU) thresholds is reported to evaluate the localization performance of our FEEL method. The IoU thresholds are set from 0.5 to 0.7 with an interval of 0.1 for the THUMOS'14 dataset, and their average mAP is adopted. For the ActivityNet v1.2 dataset, the IoU thresholds are 0.5, 0.75, and 0.95. Besides, the average mAP with IoU thresholds set from 0.5 to 0.95 with an interval of 0.05 is reported. Considering that the clustering results are crucial for our method, we adopt the conventional clustering evaluation protocol, i.e., normalized mutual information score (NMI), to validate the clustering performance of our FEEL method.  
 \begin{table}
	\caption{Localization performance comparison between our FEEL method and the state-of-the-art methods on ActivityNet v1.2 dataset. The Avg means the average mAP value with IoU thresholds set from 0.5 to 0.95 with an interval of 0.05.}
	\label{table:results on acnet}
	\centering
	\begin{tabular}{c|c|c|c|c|c}
		\hline
		\multirow{2}{*}{Supervision} & 	\multirow{2}{*}{Method} &\multicolumn{4}{c}{mAP@IoU (\%)}\\ \cline{3-6} && 0.5 & 0.75 & 0.95 & Avg\\
		\hline\hline
		\multirow{15}{*}{Weakly}
		&Clean-Net &37.1&20.3&5.0&21.6\\
		&BaS-Net & 38.5 &24.2& 5.6 &24.3\\
		&STPN &39.6& 22.5 & 4.3 &23.2\\
		&TCAM &40.0&25.0&4.6&24.6\\
		&DGAM &41.0&23.5&5.3&24.4\\ 
		&CoLA & 42.7 &25.7&5.8&26.1\\ 
		&ACSNet & 40.1 & 26.1 &6.8&26.0 \\ 	
		&CMCS&36.8&22.0 &5.6 &22.4\\
		&TSCN& 37.6 &23.7 &5.7 &23.6\\
		&HAMNet & 41.0 & 24.8 &5.3&25.1 \\ 
		&WSAL-UM & 41.2 & 25.6 &6.0&25.9 \\ 
		&AUMN & 42.0 & 25.0 &5.6&25.5 \\ 
		&UGCT&  43.1& 26.6& 6.1&26.9\\ 
		&RefineLoc& 38.7& 22.6& 5.5& 23.2\\
        &DGGNet& 44.3 & 26.9 &5.5&27.0 \\
        &AICL& 49.6 & 29.1 &5.9&29.9 \\
        &CASE& 43.8 & 27.2 &\bf{6.7}&27.9 \\ 
		&PMIL& 44.2 & 26.1 &5.3&26.5 \\ 
		\hline\hline
		\multirow{7}{*}{Unsupervised}&TCAM&35.2&21.4&3.1&21.1\\
		&STPN&28.2&16.5&3.7&16.9\\
		&WSAL-BM&28.5&17.6&\underline{4.1}&17.6\\
		&TSCN&22.3&13.6&2.1&13.6\\
		&UGCT&{37.4}&{23.8}&\bf{4.9}&{22.7}\\
		&FEEL-F&\underline{37.9}&\underline{25.4}&{3.7}&\bf{24.5}\\ 
		&FEEL-V&\bf{38.0}&\bf{25.6}&3.4&\bf{24.5}\\
		\hline
	\end{tabular}
\end{table}

\subsection{Performance Comparison (RQ1)}
\subsubsection{Baselines} 
The proposed FEEL method is compared with several unsupervised state-of-the-art methods, including  TCAM \cite{gong2020learning}, STPN \cite{nguyen2018weakly}, WSAL-BM \cite{nguyen2019weakly}, TSCN \cite{zhai2020two} and UGCT \cite{yang2022uncertainty}. For those unsupervised methods, the localization results implemented in \cite{yang2022uncertainty} are reported. Following the common settings \cite{gong2020learning}, they label the training set and then adopt all labeled videos to optimize the localization model for several iterations, and the highest performance during iterations is reported \cite{yang2022uncertainty}.
Besides, several WTAL methods are also compared, including Clean-Net \cite{liu2019weakly}, BaS-Net \cite{lee2020background}, DGAM \cite{shi2020weakly}, STPN \cite{nguyen2018weakly},  ACSNet \cite{liu2021acsnet}, TCAM \cite{gong2020learning}, CMCS \cite{liu2019completeness}, TSCN \cite{zhai2020two}, HAMNet \cite{islam2021hybrid}, AUMN \cite{luo2021action}, WSAL-UM \cite{lee2021weakly}, RefineLoc \cite{pardo2021refineloc}, DDGNet~\cite{tang2023ddg} AICL~\cite{li2023actionness}, CASE~\cite{liu2023revisiting}, PMIL~\cite{ren2023proposal}, and ISSF~\cite{yun2024weakly}.
\subsubsection{Performance Analysis}
The results of our FEEL method with two different incremental selection strategies on ActivityNet v1.2 and THUMOS'14 datasets are reported in Table \ref{table:results on acnet} and Table \ref{table:results on thumos}, where we highlight the best unsupervised results in boldface and underline the second best ones. FEEL-F and FEEL-V denote our method applies the selection strategy of constant and variable mode, respectively.

From those results, we have the following observations. For the ActivityNet v1.2 dataset, our FEEL method achieves the new unsupervised state-of-the-art localization results in terms of all metrics except for ``mAP@IoU=0.95''. Compared to the strongest UGCT baseline, the FEEL model makes great absolute improvements on the challenging ``mAP@IoU=0.75'' and ``Avg'' metrics. Moreover, it is worth noting that the FEEL method even beats several strong weakly-supervised methods like Clean-Net and BaS-Net, demonstrating the effectiveness of our incremental selection strategy.

For the THUMOS'14 dataset, except for the failure to outperform UGCT, the FEEL method consistently surpasses all other unsupervised baselines by a large margin in all metrics. The performance disparity between FEEL and UGCT on the THUMOS'14 dataset can be attributed to several factors. Firstly, the dataset contains a greater number of multi-labeled videos with extended durations, which poses a challenge in the extraction of features conducive to pseudolabeling. Furthermore, the limited quantity of training videos, coupled with a narrow range of action categories, may impede the efficacy of the self-paced learning strategy. This is due to the insufficient provision of instances necessary for establishing a robust cluster representation. However, compared to this strong competitor, our FEEL model still achieves about 1.0\% and 1.4\% improvements in ``mAP@IoU=0.6'' and ``mAP@IoU=0.7''. Furthermore, the FEEL model also outperforms several weakly-supervised methods such as AutoLoc and Clean-Net over all metrics. 

Overall, compared to the model FEEL-F of constant mode, FEEL-V yields substantial improvement on both the THUMOS'14 and ActivityNet v1.2 datasets. The excellent localization results on both datasets verify the benefits of the proposed clustering confidence improvement and two incremental selection modes.

\begin{table}
	\caption{Localization performance comparison between our FEEL method and the state-of-the-art methods on THUMOS'14 dataset. The Avg means the average mAP value with IoU threshold set from 0.5 to 0.95 with an interval of 0.05.}
	\label{table:results on thumos}
	\centering
	\begin{tabular}{c|c|c|c|c|c}
		\hline
		\multirow{2}{*}{Supervision} & 	\multirow{2}{*}{Method} &\multicolumn{4}{c}{mAP@IoU (\%)}\\ \cline{3-6} & & 0.5 & 0.6&0.7 & Avg \\
		\hline\hline
		\multirow{15}{*}{Weakly}
		&Clean-Net &23.9&13.9&7.1&15.0\\  
		&BaS-Net & 27.0& 18.6 &10.4&18.7\\  
		&STPN & 21.8 & 11.7 & 4.1&12.5\\ 
		&TCAM &30.1&19.8&10.4&20.1\\
		&DGAM &28.8&19.8&11.4&19.7\\   
		&CoLA &32.2&22.0&13.1&22.4\\	
		&ACSNet &  32.4 &22.0&11.7 &22.0\\ 
		&CMCS&23.1 &15.0 &7.0&15.0\\  
		&TSCN& 28.7 & 19.4 &10.2&19.4\\   
		&HAMNet &  31.0 & 20.7 &11.1 &20.9\\ 
		&WSAL-UM &  33.7 &22.9&12.1 &22.9\\  
		&AUMN &  33.3 &20.5&9.0&20.9 \\ 
		&UGCT& 35.8& 23.3& 11.1&23.4\\  
		&RefineLoc& 23.1& 13.3& 5.3&13.9\\ 
  	&DDGNet&41.4&27.6&14.8&27.9\\
  	&AICL&36.9&25.3&14.9&25.7\\
		&CASE&37.7&-&13.7&-\\
        &PMIL&{40.0}&\bf{27.1}&15.1&27.4\\  
        &ISSF&\bf{41.8}&{25.5}&12.8&26.7\\ 
		\hline\hline
		\multirow{7}{*}{Unsupervised}&TCAM&25.0&16.7&8.9&16.9\\
		&STPN&20.9&10.7&4.6&12.1\\
		&WSAL-BM&26.1&16.0&6.7&16.3\\
		&TSCN&26.0&15.7&6.0&15.9\\
		&UGCT&\bf{32.8}&\underline{21.6}&{10.1}&\bf{21.5}\\ 
		&FEEL-F&{28.5}&{19.7}&\underline{10.9}&{19.4}\\
		&FEEL-V&\underline{29.3}&\bf{22.6}&\bf{11.5}&\underline{20.8}\\
		\hline
	\end{tabular}
\end{table}

\begin{table*}
	\caption{Ablation studies of the proposed FEEL model on THUMOS'14 and ActivityNet v1.2 datasets where {IIS} and CCI represent the incremental selection strategy and the clustering confidence improvement, respectively. The "\checkmark" mark denotes that the corresponding module is enabled.}
	\label{Table: Ablation study}
	\centering
	\begin{tabular}{c|c|c|c|c|c|c||c|c|c|c}
		\hline
		\multirow{3}{*}{Method}&\multirow{3}{*}{IIS}&\multirow{3}{*}{CCI}& \multicolumn{4}{c||}{THUMOS'14} & \multicolumn{4}{c}{ActivityNet v1.2}\\
		\cline{4-11}
		&&& \multicolumn{3}{c|}{mAP@IoU (\%)} &\multirow{2}{*}{Avg}& \multicolumn{3}{c|}{mAP@IoU (\%)} &\multirow{2}{*}{Avg}\\  
		\cline{4-6}
		\cline{8-10}
		&&& 0.5 &0.6& 0.7 && 0.5 & 0.75 & 0.95 &\\
		\hline\hline
  		Snippet-wise UTAL &&&6.8&4.6&2.6&4.7&14.6&9.2&1.3&8.4\\
		CoLA-UTAL &&&19.6&14.2&7.3&13.7&34.1&23.2&3.4&21.5\\
        FEEL ($I_{max}=10$, w/o. IIS) &&\checkmark&23.9&16.6&9.2&16.5&37.4&25.4&3.2&24.3\\
		FEEL-F ($I_{max}=10$, w/o. CCI) &\checkmark&&23.1&16.7&9.1&16.3&37.1&25.0&3.2&24.2\\
		FEEL-V ($I_{max}=10$, w/o. CCI) &\checkmark&&25.3&17.6&9.7&17.5&36.4&25.0&3.0&23.7\\
		FEEL-F ($I_{max}=4$) &\checkmark&\checkmark&{23.5}&{16.6}&8.6&16.2&37.3&25.2&3.4&24.2\\
		FEEL-F ($I_{max}=5$) &\checkmark&\checkmark&{23.8}&{17.6}&9.8&17.1&37.5&25.5&3.3&24.4\\
		FEEL-F ($I_{max}=6$) &\checkmark&\checkmark&{24.7}&{18.1}&10.3&17.7&{37.9}&25.4&{3.7}&{24.5}\\
		FEEL-F ($I_{max}=10$) &\checkmark&\checkmark&\underline{28.5}&\underline{19.7}&\underline{10.9}&\underline{19.4}&37.6&{25.6}&{3.5}&24.4\\
  		FEEL-F ($I_{max}=15$) &\checkmark&\checkmark&{24.4}&{17.4}&{9.4}&{17.0}&{37.7}&{25.6}&\bf{4.1}&24.5\\
  		FEEL-F ($I_{max}=20$)&\checkmark&\checkmark&{24.1}&{16.0}&8.3&16.5&\underline{38.1}&\bf{25.9}&{3.5}&\underline{24.7}\\
		FEEL-V ($I_{max}=4$)&\checkmark&\checkmark&{24.2}&{17.3}&{9.7}&{17.1}&{37.2}&{25.5}&{3.2}&24.3\\
		FEEL-V ($I_{max}=5$)&\checkmark&\checkmark&{25.6}&{18.2}&{10.2}&{18.0}&{37.5}&{25.5}&{3.3}&{24.4}\\
		FEEL-V ($I_{max}=6$) &\checkmark&\checkmark&{26.2}&{18.7}&{10.3}&{18.4}&{37.3}&{25.4}&{3.1}&{24.4}\\
		FEEL-V ($I_{max}=10$) &\checkmark&\checkmark&\bf{29.3}&\bf{22.6}&\bf{11.5}&\bf{20.8}&{38.0}&{25.6}&3.4&{24.5}\\
		FEEL-V ($I_{max}=15$)&\checkmark&\checkmark&{23.8}&{16.5}&{9.7}&{16.7}&{38.1}&\underline{25.7}&3.3&{24.6}\\
		FEEL-V ($I_{max}=20$) &\checkmark&\checkmark&{23.4}&{15.8}&{8.1}&{15.9}&\bf{38.4}&{25.6}&\underline{3.9}&\bf{25.0}\\
		\hline
	\end{tabular}
\end{table*}

\subsection{Ablation Study (RQ2)}
In this section, a series of ablation studies have been conducted on the THUMOS'14 and ActivityNet v1.2 datasets to look deeper into the effectiveness of different components in our FEEL model, including the CCI module and incremental selection. Particularly, we generate the following model variants by eliminating one or two modules at a time. 
\begin{itemize}
	\vspace{-2pt}
 	\item FEEL-F ($I_{max}=10$, w/o. IIS): We remove the IIS module from our full model, i.e., clustering all videos with the CCI module and then directly trains the localization model with the entire pseudolabeled dataset with the total iteration $I_{max}$ setting to 10.
	\item FEEL-F ($I_{max}=10$, w/o. CCI) or FEEL-V ($I_{max}=10$, w/o. CCI): We remove the CCI module from our full model, and only adopt two different incremental selection strategies with the total iteration $I_{max}$ setting to 10.
	\item CoLA-UTAL: We discard both the CCI module and incremental selection from our full model. Specifically, during each iteration, the training videos are labeled based on the Euclidean clustering results and then the entire labeled set is used to train the CoLA model, where the highest localization performance is reported. 
 	\item Snippet-wise UTAL:   we extracted the features of top-$k$ action-positive snippets in each video, resulting in $k*N$ snippets, which are then subjected to clustering. The pseudolabels of a video depend on the clustering results of its corresponding $k$ snippets.
\end{itemize}

The ablation results are listed in Table \ref{Table: Ablation study}, where we mark the adopted module with a ``\checkmark '' symbol. From those results, the following conclusions stand out:
\begin{itemize}
	\vspace{-2pt}
	\item  Firstly, as the iteration number $I_{max}$ increases, the localization results of both the FEEL-F and FEEL-V method demonstrate a gradual growth trend, which is expected because a larger  $I_{max}$  implies a more stable training data growth and higher sample quality per iteration. As we can observe, enhancing $I_{max}$ larger than 10 still improves performance on ActivityNet v1.2, a large dataset with 100 categories, due to the more gradual instance selection. However, this does not apply to the smaller THUMOS'14 dataset with 20 categories, where selecting too few instances initially can harm learning due to poor class representation. Thereafter, We have determined that setting $I_{max}$ to ensure that the selected instances in the first iteration at least reaches or exceeds the number of action categories K (which is a known annotation), is a practical selection guideline, and setting the $I_{max}$ to 10 strikes a balance between the two datasets.

	\item Secondly, analyzing the performance of FEEL ($I_{max}=10$, w/o. CCI), FEEL ($I_{max}=10$, w/o. IIS) and FEEL ($I_{max}=10$) with different selection modes together, we find that if the CCI or IIS module is removed, our FEEL model suffers from great performance drop over all metrics, especially on THUMOS'14 dataset, which verifies that the synergistic effects of CCI and IIS module contribute to the pseudolabel qualities significantly.
	
	\item Finally, when compared with the CoLA-UTAL, both FEEL ($I_{max}=10$, w/o. CCI) variants achieve significant improvements over all metrics on both datasets, which indicate that both proposed incremental selection strategies substantially enhance the model learning. Moreover, the results of Snippet-wise UTAL, of which the NMI is 51.0\% on Thumos'14, are even inferior to CoLA-UTAL baseline. These results verify that the snippet-wise clustering is unadvisable, since it is hard to distinguish the classes of all snippets.
\end{itemize}

\begin{figure}[t!]
	\centering
	\label{parameterimg}
	\subfigure[Clustering results w.r.t iterations.]{
		\label{fig:ablation_iteartion:a} 
		\resizebox{.44\columnwidth}!{
			\includegraphics{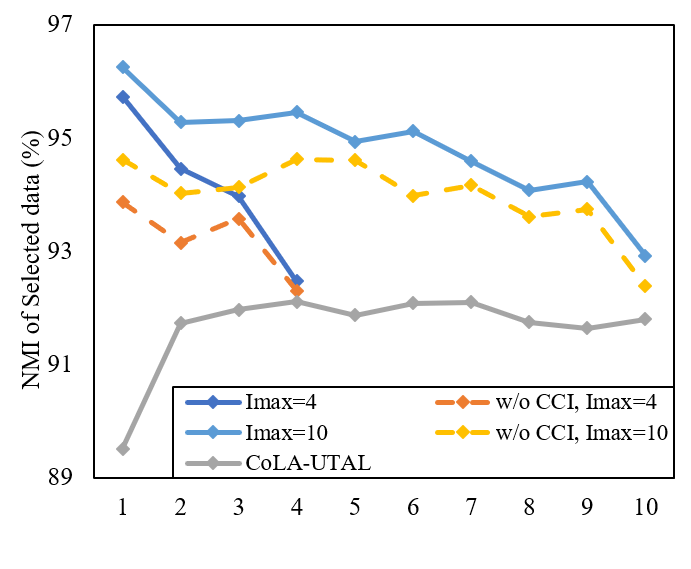}
	}}
	\quad
	\subfigure[Localization results w.r.t iterations.]{
		\label{fig:ablation_iteartion:b} 
		\resizebox{.44\columnwidth}!{
			\includegraphics{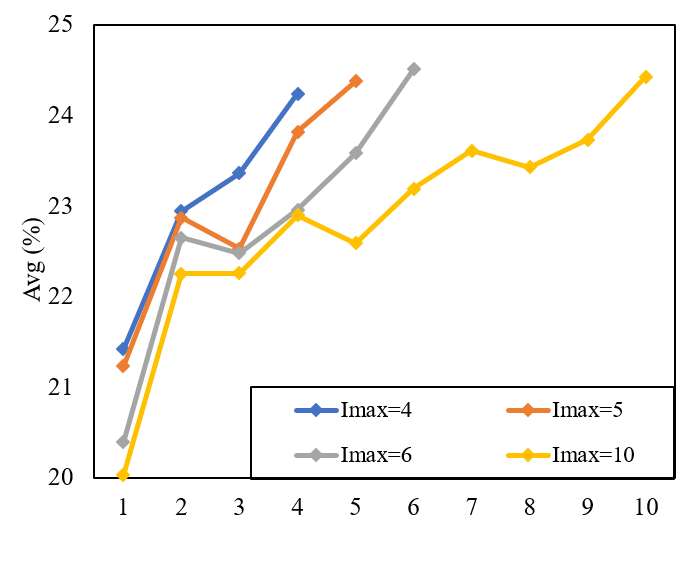}
	}}
	\caption{Localization results and clustering results of our FEEL-F model w.r.t iterations.} 
\end{figure}


\subsection{Influence of $I_{max}$ on model performance (RQ3)}

\subsubsection{Clustering Performance w.r.t iterations with different $I_{max}$.} Fig \ref{fig:ablation_iteartion:a} shows the NMI results of the selected videos among different iterations on ActivityNet v1.2 dataset. The NMI results of FEEL-F (w/o. CCI) variants and CoLA baseline are also provided. Specifically, the NMI results of our FEEL model with larger $I_{max}$ beats that of the FEEL model with smaller $I_{max}$, which indicates a more stable enlarging speed brings higher reliability of selected videos. Moreover, our full FEEL-F model consistently outperforms the corresponding FEEL-F (w/o. CCI) variant during the iterations, and all the variants of FEEL-F model achieve significant improvements compared to the CoLA baseline. All the above results confirm the effectiveness of our method to improve labeling quality.

\subsubsection{Localization Performance w.r.t iterations with different $I_{max}$.} Fig \ref{fig:ablation_iteartion:b} compares the ``Avg'' results of the FEEL-F model among different iterations on ActivityNet v1.2 dataset. From this figure, it can be seen that the best localization results of FEEL-F with different $I_{max}$ are achieved after the final iteration, because the training data reaches its maximum and the overall labeling accuracy is very high in the last iteration. Besides, although the localization results are decreased after some iterations, the localization results of FEEL-F model with different $I_{max}$ show an increasing trend with the iterations. 

\begin{figure}[!t]
	\centering
	\includegraphics[width=0.85\columnwidth]{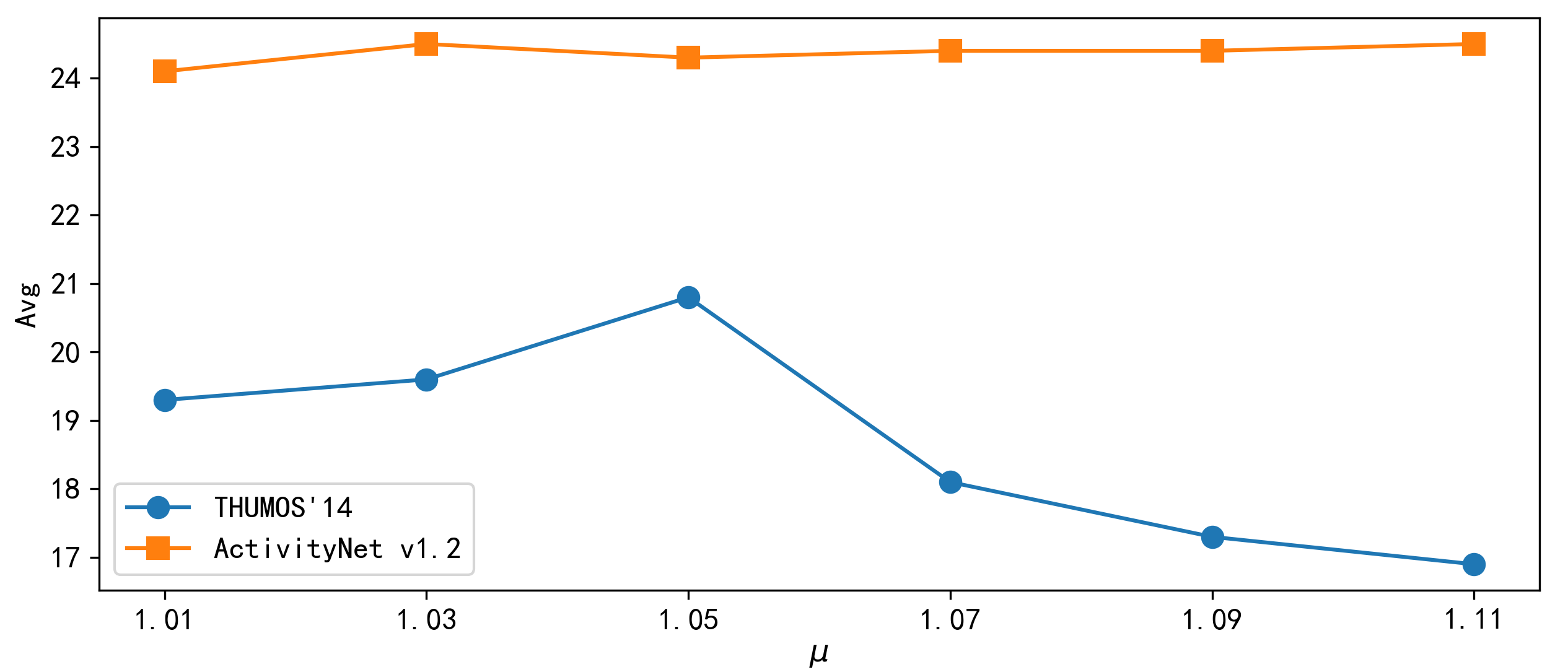}
	\caption{The ``Avg'' Localization results of our FEEL-V model w.r.t different $\mu$. $I_{max}$ is set to 10.}
	\label{fig:ablation_mu}
\end{figure}
 
\subsubsection{Localization Performance w.r.t different $\mu$.} Fig \ref{fig:ablation_mu} compares the ``Avg'' results of the FEEL-V model among different $\mu$ on both datasets. Specifically, the performance on the ActivityNet v1.2 dataset remains robust across a variety of $\mu$ values, suggesting a higher tolerance for different selection rates. However, for the THUMOS dataset, where the initial number of training samples is limited, there is a notable decline in performance when $\mu$ exceeds 1.05. As we have claimed, this THUMOS dataset, which contains a relatively small number of videos across 20 classes, requires a balance to ensure that the initial selection of training instances adequately maintains class representation.

\subsection{Scalability on UTAL (RQ4)}

To assess the scalability of our FEEL model, we integrated the CCI and IIS components, with the constant mode FEEL-F and the iteration number $I_{max}=4$, into the TCAM and STPN UTAL methods. This led to the enhanced models FEEL-F ($I_{max}=4$) + TCAM and FEEL ($I_{max}=4$) + STPN. We also adapted the weakly-supervised AICL and CASE methods for unsupervised settings by employing K-means for pseudolabeling and then applied FEEL to these models, and then employed CCI and IIS modules for evaluations.
The results, as shown in Table \ref{table:results with different baselines}, indicate that our enhancements led to significant performance improvements across all metrics for the baseline models, confirming FEEL's scalability. Notably, AICL, CASE, and the COLA we adopted show greater improvements due to their use of the contrastive learning paradigm within the video for distinguishing action snippets, which is crucial for UTAL's global video feature generation.




\begin{table*}
	\caption{Localization performance comparison between original and improved UTAL models on THUMOS'14 and ActivityNet v1.2 dataset.}
	\centering
	\label{table:results with different baselines}
	\begin{tabular}{c|c|c|c|c||c|c|c|c}
				\hline
		\multirow{3}{*}{Method}&\multicolumn{4}{c||}{THUMOS'14} & \multicolumn{4}{c}{ActivityNet v1.2}\\
		\cline{2-9}
		& \multicolumn{3}{c|}{mAP@IoU (\%)} &\multirow{2}{*}{Avg}& \multicolumn{3}{c|}{mAP@IoU (\%)} &\multirow{2}{*}{Avg}\\  
		\cline{2-4}
		\cline{6-8}
		& 0.5 &0.6& 0.7 && 0.5 & 0.75 & 0.95 &\\
		\hline\hline
		TCAM&25.0&16.7&8.9&16.9&35.2&21.4&3.1&21.1\\
		FEEL-F ($I_{max}=4$) + TCAM&26.1&18.3&10.5&18.3&{36.2}&{23.1}&3.4&{23.5}\\
		STPN&20.9&10.7&4.6&12.1&28.2&16.5&{3.7}&16.9\\
		FEEL-F ($I_{max}=4$) + STPN&21.7&11.0&6.9&13.2&31.1&18.3&3.5&19.6\\
        CASE\textsuperscript{*}&28.5&17.9&10.1&18.8&38.2 & 26.0 &5.6&25.7\\
		FEEL-F ($I_{max}=4$) + CASE\textsuperscript{*}&31.3&19.6&11.5&20.8&41.9&28.2&6.0&27.4\\
		AICL\textsuperscript{*}&29.7&19.8&11.1&20.2&44.3&27.8&6.2&28.1\\
		FEEL-F ($I_{max}=4$) + AICL\textsuperscript{*}&33.2&20.7&12.1&22.0&46.1&28.7&6.3&28.7\\
		\hline

	\end{tabular}
\end{table*}

\begin{figure}[t!]
	\centering
	\subfigure[The ``wrapping presents'' result]{
		\label{fig:exp_visual:a} 
		\includegraphics[width=0.90\columnwidth]{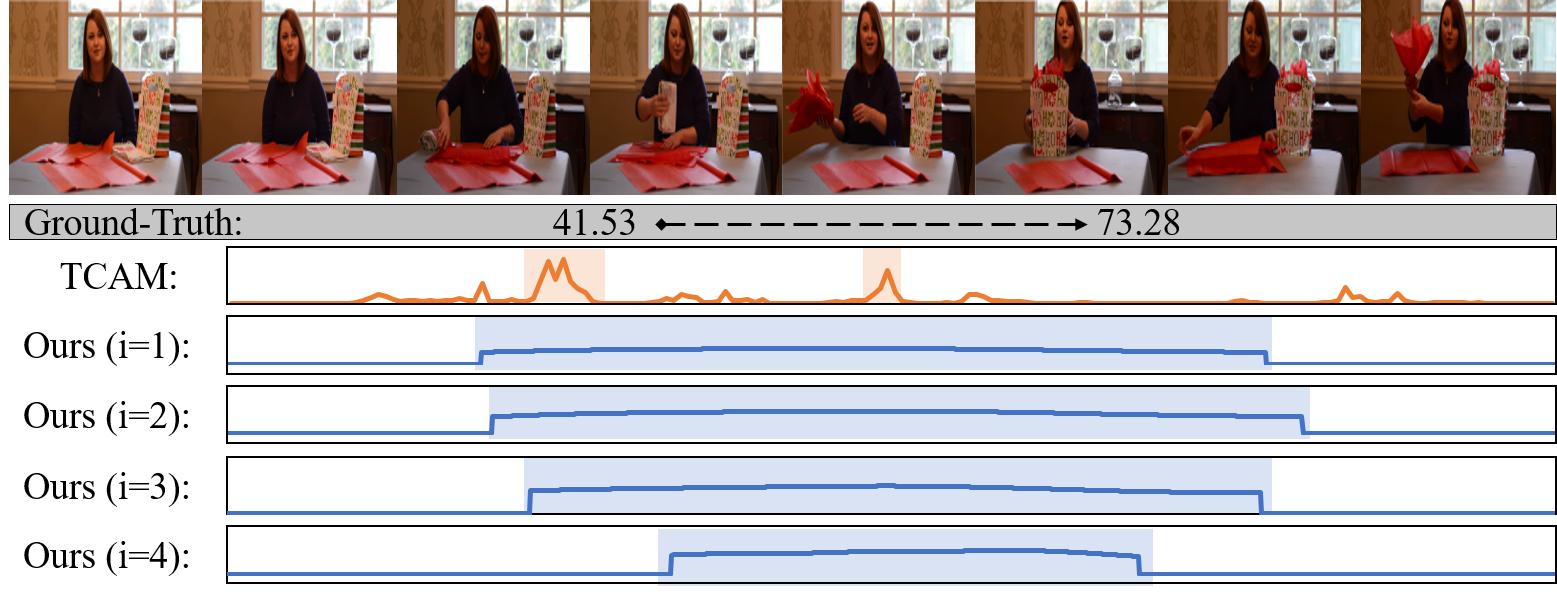}}
	\centering
	\subfigure[The ``preparing the pasta'' result]{
		\label{fig:exp_visual:b} 
		\includegraphics[width=0.90\columnwidth]{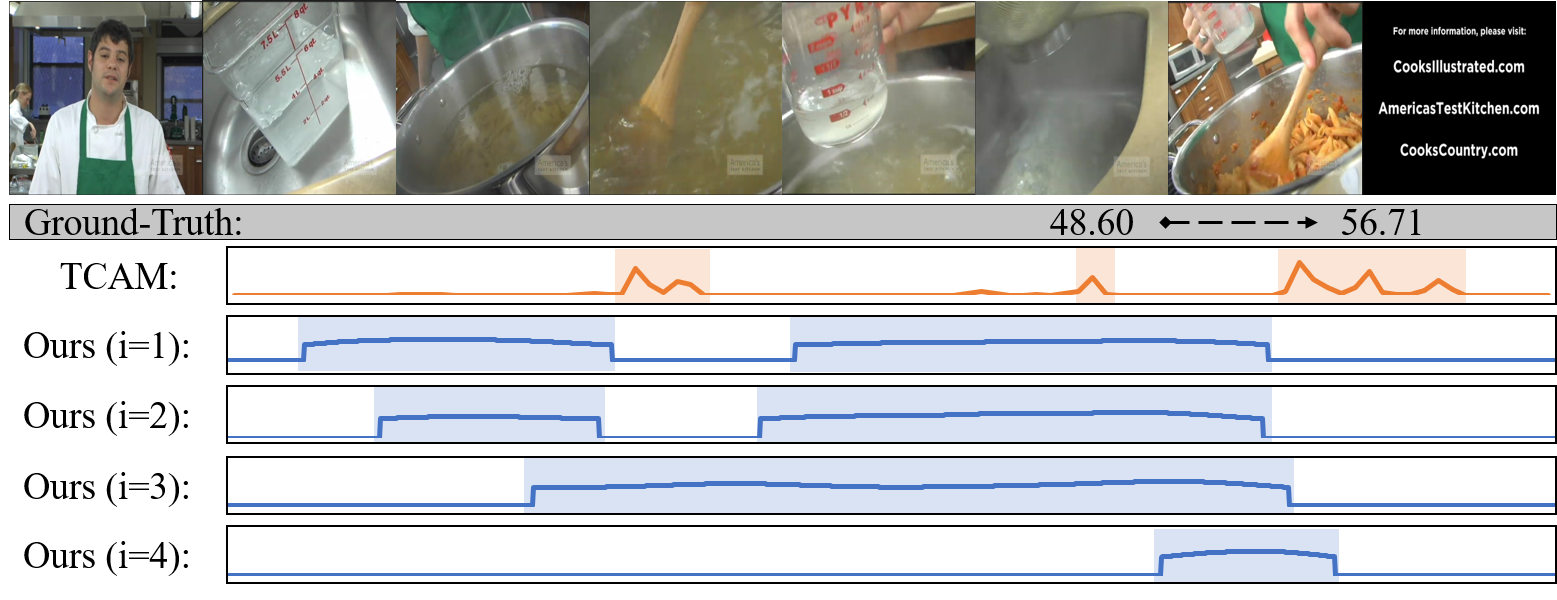}
	}
	\caption{Qualitative localization results by TCAM and our methods on ActivityNet v1.2 dataset. Our model of constant mode FEEL-F ($I_{max}=4$) is adopted, where i=1, 2, 3, 4 denotes the localization result after the corresponding iteration.}
	\vspace{-2pt}
	\label{fig:exp_visual}
\end{figure}

\subsection{Qualitative Visualization}
We provide the visualization results of two actions in the ActivityNet v1.2 dataset. The predicted class-agnostic attention weight of the TCAM model during its iterations and the FEEL-F ($I_{max}=4$) model in four iterations are also presented in Fig \ref{fig:exp_visual}. As we can see, the TCAM model returns several short and sparse intervals slightly overlapping the desired video proposal in these two action cases. In contrast, our FEEL-F model after the first iteration can already return a satisfying result. As the iteration goes on, the localization model becomes stronger and the results are enhanced gradually for both actions. After the final iteration, the FEEL-F model successfully returns the entire desired proposal with the highest IoU performance for those two cases. From those visualization results, we can find that the proposed CCI module and incremental selection strategy can collaboratively improve the quality of training data and thus greatly improve the localization performance during the iteration process. 

\begin{figure}[t!]
	\centering
	\subfigure[T-SNE after intial clustering.]{
		\label{fig:clusteringvisualization:a} 
		\resizebox{.44\columnwidth}!{
			\includegraphics{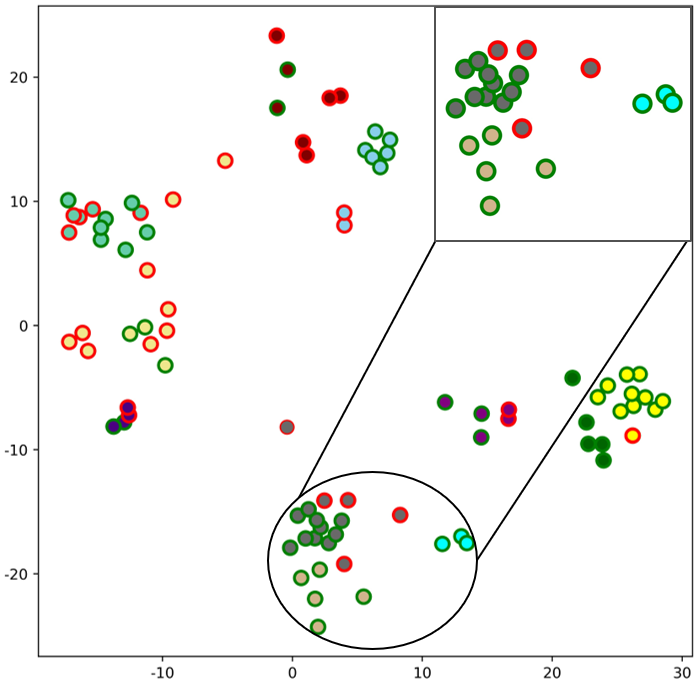}
	}}
	\quad
	\subfigure[T-SNE after the FEEL process.]{
		\label{fig:clusteringvisualization:b} 
		\resizebox{.44\columnwidth}!{
			\includegraphics{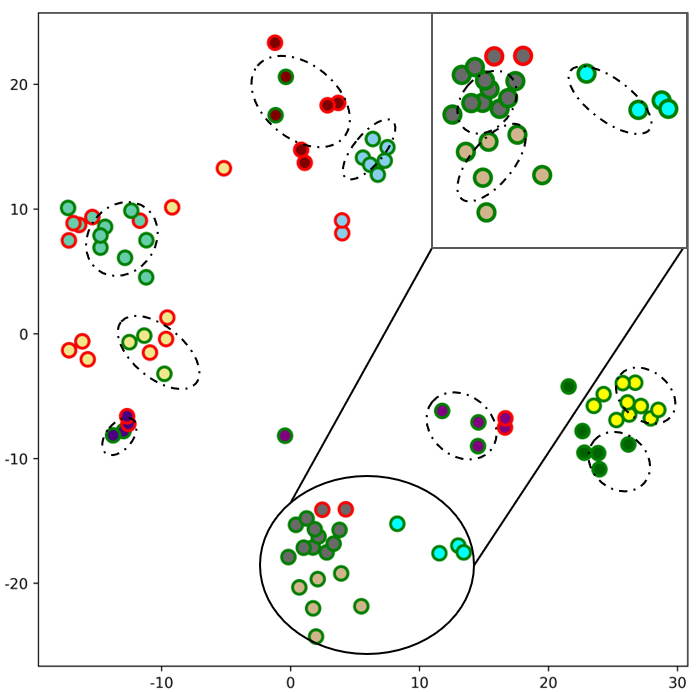}
	}}
	\caption{Visualization of the CCI module and incremental selection. The green borderline for a dot means a correct pseudolabel, while the red borderline means the opposite.} 
	\label{fig:clusteringvisualization}
\end{figure}

Fig \ref{fig:clusteringvisualization} shows the T-SNE \cite{van2008visualizing} visualization of how our FEEL model works. In this figure, the green borderline for a dot means a correct pseudolabel, while the red borderline means the opposite. Compared to the T-SNE result of the initial clustering in Fig \ref{fig:clusteringvisualization:a}, the result in Fig \ref{fig:clusteringvisualization:b} indicates that our CCI module corrects many mislabeled dots with the red borderline into the correct ones with the green borderline, and encourages the correctly labeled videos to be top-ranked in their corresponding clusters as well. As shown in the right top of Fig \ref{fig:clusteringvisualization:b}, our incremental selection strategy samples a small portion of those top-ranked videos from each cluster after the label correction, resulting in an even higher quality of video annotations for the subsequent model training.

\section{Conclusion and Future work}
To address the unsupervised temporal action localization, we present a self-paced iterative learning model FEEL. It is the first effort to address UTAL with the self-paced learning paradigm. To improve the generation quality of the pseudolabeled videos, we introduce a clustering confidence improvement module, which utilizes the feature-robust Jaccard distance to refine the original video clustering results and improve label prediction capability. Moreover, we present a self-paced incremental instance selection, which is able to automatically choose an increasing portion of the most reliable pseudolabeled videos for easy-to-hard localization model training. Extensive experiments, ablation studies, hyper-parameter analysis, and visualization qualitative results have well-verified the effectiveness of our model.

In the future, aiming for continuous exploration of UTAL, we intend to integrate the mutual learning mechanism and multi-modal pretraining network into our model training, thereby improving the overall localization performance.

\bibliographystyle{ACM-Reference-Format}

\bibliography{sample-manuscript}

\end{document}